\documentclass{article}

    \PassOptionsToPackage{numbers, compress}{natbib}

 \usepackage[creativeai, final]{neurips_2025}
\usepackage{graphicx}

\usepackage{xcolor}

\usepackage{xcolor}
\usepackage{soul}
\usepackage{listings}

\definecolor{thoughtcolor}{gray}{0.90}

\newcommand{\aithought}[1]{{\sethlcolor{thoughtcolor}\hl{\texttt{#1}}}}

\usepackage[utf8]{inputenc} 
\usepackage[T1]{fontenc}    
\usepackage{hyperref}       
\usepackage{url}            
\usepackage{booktabs}       
\usepackage{amsfonts}       
\usepackage{nicefrac}       
\usepackage{microtype}      
\usepackage{xcolor}         

\title{Semantic Glitch: Agency and Artistry in an Autonomous Pixel Cloud}
\author{
  Qing Zhang \\ The University of Tokyo  \\ \href{mailto:qzkiyoshi@gmail.com}{qzkiyoshi@gmail.com}, \And
  Jing Huang \\ Tokyo University of the Arts \\ \href{mailto:hkoukenj@gmail.com}{hkoukenj@gmail.com}, \And
  Mingyang Xu \\ Keio University \\ \href{mailto:mingyang@kmd.keio.ac.jp}{mingyang@kmd.keio.ac.jp}, \And
  Jun Rekimoto \\ The University of Tokyo \\SONY CSL Kyoto\\ \href{mailto:rekimoto@acm.org}{rekimoto@acm.org} \\
}

\begin{document}

\maketitle

\begin{abstract}
    While mainstream robotics pursues metric precision and flawless performance, this paper explores the creative potential of a deliberately ``lo-fi'' approach. We present the ``Semantic Glitch,'' a soft flying robotic art installation whose physical form—a 3D pixel style cloud—is a ``physical glitch'' derived from digital archaeology. We detail a novel autonomous pipeline that rejects conventional sensors like LiDAR and SLAM, relying solely on the qualitative, semantic understanding of a Multimodal Large Language Model to navigate. By authoring a bio-inspired personality for the robot through a natural language prompt, we create a ``narrative mind'' that complements the ``weak,'' historically-loaded body. Our analysis begins with a 13-minute autonomous flight log, and a follow-up study statistically validates the framework's robustness for authoring quantifiably distinct personas. The combined analysis reveals emergent behaviors—from landmark-based navigation to a compelling ``plan-to-execution'' gap—and a character whose unpredictable, plausible behavior stems from a lack of precise proprioception. This demonstrates a lo-fi framework for creating imperfect companions whose success is measured in character over efficiency.

\end{abstract}

\section{Introduction}
In an era where digital imagery relentlessly pursues high fidelity, why has the ``pixel'' aesthetic, born from technical limitations, sparked a persistent wave of retro-futurism? \cite{menkman2011glitch, hertz2012zombie} Furthermore, when a symbol composed of pixels, which should exist on a two-dimensional screen, suddenly acquires a physical body and floats among us like a seemingly autonomous creature, how does our relationship with it, and our perception of the virtual and the real, change? \cite{kac2005telepresence, chen2025multi} This paper explores these questions by detailing the creation and behavior of the ``Pixel Cloud,'' a soft robotic art installation that gains its physical autonomy from a Multimodal Large Language Model (MLLM). Our approach to authoring an agent's character builds on artistic and scientific explorations into crafting lifelike \cite{lachenmyer2022aquarium, lachenmyer2022b}, emergent behaviors for interactive robotic agents. This work does not aim to solve any practical problem. Instead, it follows the ``Speculative Design'' philosophy advocated by Anthony Dunne and Fiona Raby \cite{dunne2013speculative}, functioning as a ``speculative object'' to provoke public imagination and debate. It poses a series of ``what if'' questions: What if the untouchable digital ``cloud'' had a visible, fragile, physical body? What if the symbols from our digital childhood memories gained physical autonomy? By combining media archaeology \cite{hertz2012zombie} with robotics, the core thesis of this work is that through a ``physical hack'' \cite{zareei2015physical} of the pixel, we can reveal and reshape the increasingly complex ``entangled agencies'' \cite{barad2007meeting} among humans, machines, and the environment.

Grounded in media archaeology and speculative design \cite{auger2012robot, keane2017human}, this paper details the symbiotic creation of the ``Pixel Cloud'' from its ``physical glitch'' body to its narrative Al mind. We then analyze an autonomous flight as a deep case study to demonstrate how our novel two-stage pipeline fosters emergent, goal-oriented behaviors. Critically, to address the limitations of a single case study, we then present an expanded validation that confirms our ability to author multiple, statistically distinct personas. We conclude by discussing the implications of this ``lo-fi'' approach and our vision for creating more relatable machine companions.

\section{The Body: A Deliberate ``Physical Glitch''}

\begin{figure}
    \centering
    \includegraphics[width=\linewidth]{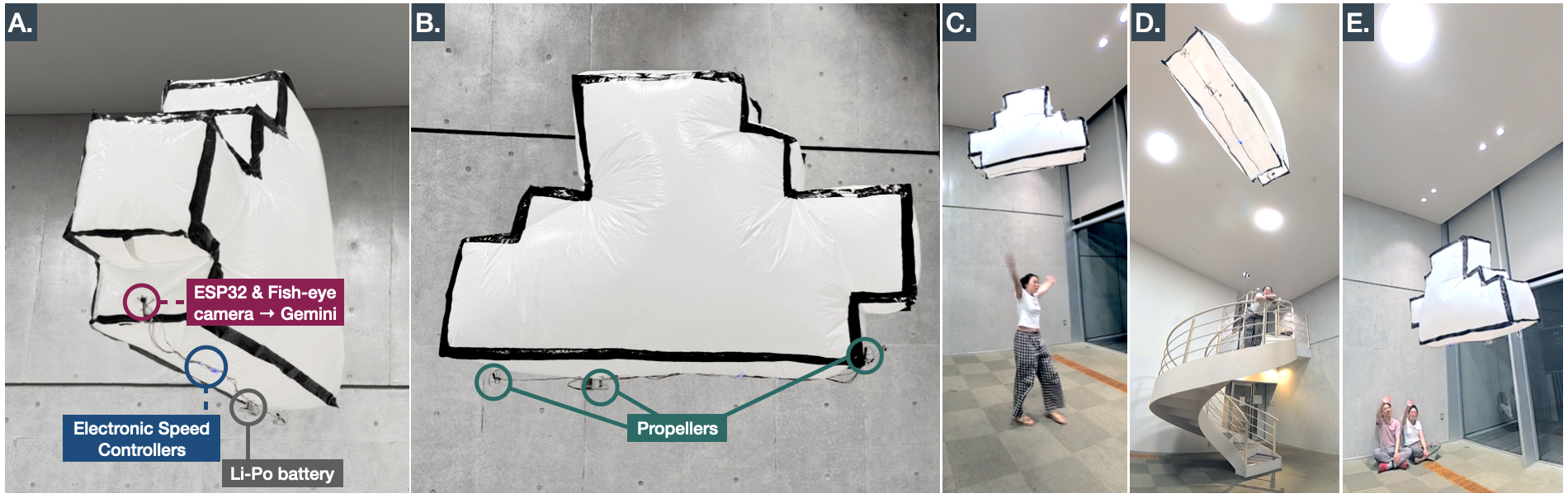}
    \caption{The ``Semantic Glitch'' hardware, flight behavior, and first-person perspective with MLLM-generated reasoning. (A, B) The robot's body, showing the placement of the ESP32 with a fish-eye camera, electronic speed controllers, Li-Po battery, and propeller modules. (C, D, E) The robot in flight, demonstrating its interaction with the environment and its ``perspective-dependent morphological illusion.''}
    \label{fig:system_design}
\end{figure}

The robot's physical form is a deliberate ``physical glitch,'' designed to embody the ``Yowai Robotto'' (Weak Robot) philosophy by rejecting metric precision in favor of character \cite{nowacka_diri_2015, xu2025cuddle, yamada_zerone_2019, okada2022weak}. A core engineered feature is its ``perspective-dependent morphological illusion'': from one angle, it appears as a 2D pixel image, but as it rotates, its 3D voxel structure is revealed (Fig. \ref{fig:system_design} C-E). This effect translates a software ``error'' into a tangible, repeatable imperfection. Constructed as a soft, fragile helium blimp, its form is intentionally ``weak'' to invite empathetic interaction \cite{okada2022weak, flocchini1999hard}. This physical weakness is the direct counterpart to the agent's cognitive framework, which, as we will show, lacks precise physical self-awareness (proprioception). This mismatch between a high-level semantic mind and a low-fidelity body creates the emergent, non-optimal behaviors at the core of our work.

\section{The Mind: Navigation as Bio-Inspired Narrative}

\textbf{Rejecting Metric Precision: }The conventional path to robotic autonomy involves building a precise, mathematical model of the world. This is typically achieved with a suite of metric sensors (such as LiDAR or Infrared Depth Sensor) and complex algorithms like SLAM (Simultaneous Localization and Mapping), a technology envisioned as a future step in the project's initial conceptualization. We deliberately rejected this path. A SLAM-based robot, with its metric geometric understanding, would be philosophically out of character.'' Its calculated, optimal movements would be incongruous with the artifact's ephemeral nature, breaking the illusion of an animate entity. Therefore, to maintain the weak robot'' concept, the mind's perception had to be as abstract as the physical form.

\begin{figure}
    \centering
    \includegraphics[width=\linewidth]{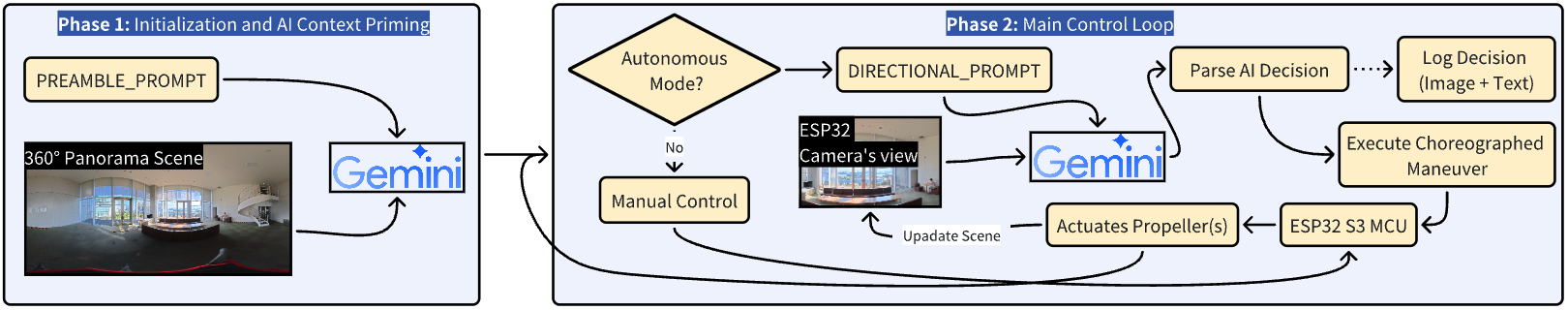}
    \caption{The two-phase semantic reasoning pipeline. Phase 1 (Initialization): A single 360$^{\circ}$ panorama and a PREAMBLE\_PROMPT are sent to the Gemini API to establish a stateful ``mental map''. Phase 2 (Control Loop): In a continuous loop, the system uses the live camera view and a DIRECTIONAL\_PROMPT to generate context-aware actions, which are then logged and executed by the robot.}
    \label{fig:workflow}
\end{figure}

\textbf{The ``Lo-Fi'' Semantic Engine: }In place of a complex, sensor-heavy system, we embraced a framework of stateful semantic reasoning. The agent's autonomy is powered by a novel, two-stage ``lo-fi'' pipeline that separates global scene understanding from local decision-making.

The entire control loop is orchestrated by a host computer (MacBook Pro, M4 Max, 64GB RAM) running a Python script, which communicates with the robot's XIAO ESP32S3 core. The ESP32S3 is responsible only for low-level tasks: streaming video from its camera (160$^{\circ}$ fish eye lens) and actuating its propellers via WebSocket commands. All cognition is offloaded to the remote Gemini 2.5 FLASH API, subject to its terms of use, transforming it into a stateful, MLLM ``mind.'' 




This is achieved through two distinct phases, as illustrated in Figure \ref{fig:workflow}. First, the \textbf{Preamble Stage} performs zero-shot spatial mapping. Upon initialization, the system begins a stateful ChatSession with the Gemini API. It sends the PREAMBLE\_PROMPT along with a single $360^{\circ}$ panoramic image of the environment. This one-time action tasks the Al with performing a high-level analysis of the entire operational area, identifying boundaries, major landmarks, open fly-zones, and obstacles before any physical movement occurs. This establishes a persistent ``mental map'' within the Al's chat context, a process that took 2.81 seconds to complete during our experiment.

Second, the \textbf{Directional Stage} handles context-aware deliberation. In the main operational loop, we recast navigation as a continuous Visual Question Answering (VQA) problem \cite{antol2015vqa}. For each decision, the DIRECTIONAL\_PROMPT is posed as the ``question,'' which the MLLM ``answers'' by interpreting the ``vision'' context from the ESP32S3's live video frame. The resulting answer, containing both a command and a narrative reason, is informed by the global spatial map established in the preamble stage. This creates a continuous, state-aware feedback loop where local perception is fused with global memory. During operation, this decision loop exhibited a mean latency of $2.8 \pm 0.3$ seconds, quantitatively defining the agent's deliberate, non-continuous cognitive cycle.

\textbf{Prompt Engineering for Hierarchical Cognition: }The agent's hierarchical cognitive process is authored entirely through two carefully engineered natural language prompts. These prompts define the function of each stage of the reasoning pipeline.

The PREAMBLE\_PROMPT serves as the high-level cartographer, instructing the model to deconstruct the panoramic scene into a structured, semantic map. The DIRECTIONAL\_PROMPT acts as the low-level navigator. It instructs the agent to use its established ``mental map'' as prior knowledge when interpreting the live camera feed to make an immediate choice. The prompt explicitly defines the agent's available actions; the full, verbatim text for both the PREAMBLE\_PROMPT and DIRECTIONAL\_PROMPT is provided in Appendix \ref{sec:appendix}.




This two-prompt structure fundamentally changes the nature of navigation. The AI does not simply react to pixels or hard-coded logics; it situates its local perception within a persistent, global narrative, allowing for more sophisticated, long-term reasoning without the overhead of conventional mapping algorithms. Ultimately, this pipeline serves as a new model for authoring complex Al behavior, where high-level strategic goals and low-level personality traits can be defined and layered through natural language \cite{zitkovich2023rt}. By representing all outputs as text, our work aligns with a broader trend of using language as a unified interface for robotic control, though we apply it here to generate character-rich narrative instead of precise coordinates.

\begin{figure}
    \centering
    \includegraphics[width=\linewidth]{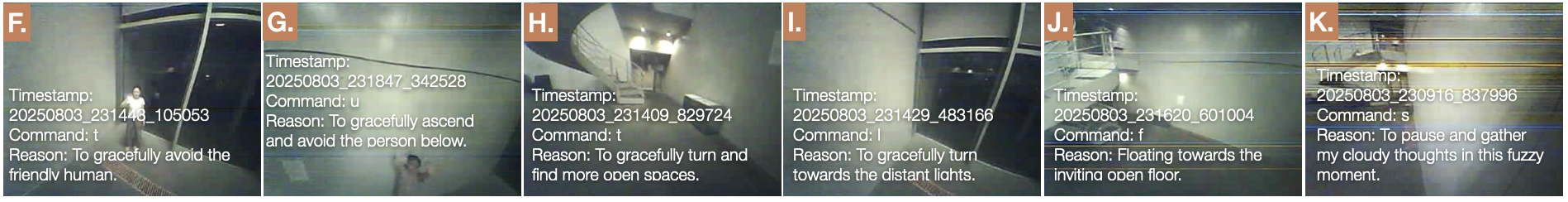}
    \caption{Key moments from the agent's first-person perspective, linking its visual input to its logged decisions. (F) Lateral avoidance of a person. (G) Vertical avoidance of a person. (H) A corrective turn near the staircase, illustrating the plan-to-execution gap. (I) Goal-oriented navigation towards distant lights. (J) Seeking open space. (K) A moment of contemplative inaction.}
    \label{fig:cloud_design_perspective}
\end{figure}

\section{Analysis of Emergent Dialogue: A Case Study in Stateful Navigation}

This section analyzes a 13-minute continuous operational log to demonstrate how the synthesis of a lo-fi form and a stateful mind produces a uniquely plausible form of artificial life, using examples illustrated in Figure \ref{fig:cloud_design_perspective}.

\textbf{Goal-Oriented Navigation and Landmark Use:} The logs confirm that the Preamble stage was successful. The agent consistently uses landmarks identified in the 360$^{\circ}$ panorama to inform its navigation, demonstrating effective use of its ``mental map.'' This is evident in decisions where it actively seeks out desirable zones, such as when turning with the reason \aithought{\texttt{``t, To gracefully turn towards the distant lights''}} (Fig. \ref{fig:cloud_design_perspective}-I) or moving forward because it is \aithought{\texttt{``f, Floating towards the inviting open floor''}} (Fig. \ref{fig:cloud_design_perspective}-J). The spiral staircase was a particularly prominent landmark, with the agent often reasoning about it as a point of interest (e.g., \aithought{\texttt{``l, To drift away from the wall and admire the elegant spiral''}}). This behavior shows the agent is not just reactively avoiding obstacles but is executing long-term, goal-oriented exploration based on the initial context.

\textbf{Emergent Social Robotics and Character:} The ``bio-inspired'' personality authored in the prompts resulted in sophisticated social behaviors. The agent recognized and reacted to humans in a way that was consistent with its prompt-defined nature. All interactions involving human observers were conducted under a protocol approved by our university's Institutional Review Board. \textbf{Dynamic Human Avoidance}: The agent employed varied strategies when encountering people. On one occasion, it chose a lateral maneuver: \aithought{\texttt{``t, To gracefully avoid the friendly human''}} (Fig. \ref{fig:cloud_design_perspective}-F). In a different situation, it opted for a vertical solution: \aithought{\texttt{``u, To gracefully ascend and avoid the person below''}} (Fig. \ref{fig:cloud_design_perspective}-G). This demonstrates a flexible decision-making capability, choosing different actions for similar problems based on the specific context. \textbf{Contemplative Behavior}: The agent's character was further revealed in moments of inaction. The log contains entries like \aithought{\texttt{``s, To pause and gather my cloudy thoughts in this fuzzy moment''}} (Fig. \ref{fig:cloud_design_perspective}-K). These are not error states; they are authored behaviors that give the agent a plausible, non-utilitarian, and creature-like quality, directly supporting the ``Yowai Robotto'' concept.

\textbf{The Plan-to-Execution Gap and the ``Yowai Robotto'':} The most notable ``glitchy'' behaviors arise from the conflict between the agent's high-level semantic understanding and its lack of low-level physical self-awareness. The agent knows what it wants to do but not precisely how to do it. For instance, after successfully navigating near the staircase, it decides to make a corrective turn with the command \aithought{\texttt{``t, To gracefully turn and find more open spaces''}} (Fig. \ref{fig:cloud_design_perspective}-H). The image shows it is close to the structure, and while the intention is correct, the subsequent maneuver is clumsy. It lacks the precise proprioceptive knowledge of its own momentum or the physical dynamics of its forward-arcing maneuver. This constant ``wrestling'' with its own physical form—a direct result of the plan-to-execution gap—is a clear and authentic expression of a ``weak'' agent that feels organic rather than programmed.

\textbf{The Voice of the Glitch: Narrative as an Artistic Medium:} Beyond the dialogue expressed through physical movement, the ``Semantic Glitch'' communicates through a third, crucial modality: its own generated text. The simple string of text that accompanies each action—such as \aithought{\texttt{``l, To drift away from the wall and admire the elegant spiral''}} or \aithought{\texttt{``s, To ponder the shimmering, uncertain view''}}—functions as more than a mere debug log or explanation. It is a performance of an ``internal monologue,'' a textual broadcast from the agent's narrative core. This output can be contrasted with the functional ``Chain-of-Thought'' (CoT) \cite{wei2022chain} reasoning used in performance-oriented driving models to generate an explicit driving rationale for improved safety and accuracy. In our work, this textual output is not a means to a more accurate end, but a distinct artistic medium in itself, what can be interpreted as a form of minimalist, AI-driven poetry that completes the artifact's character.

The poetics of this voice are ``lo-fi'' by design, mirroring the aesthetics of the body and mind. The phrases are short, direct, and laden with qualitative, emotional language. Rather than using simple objective descriptions, the agent's vocabulary is rich with evocative verbs like ``admire,'' ``embrace,'' ``pirouette,'' and ``ponder.'' Similarly, the moments of contemplative hesitation, such as \aithought{\texttt{``s, To pause and gather my cloudy thoughts...''}}, are a potent performance of the vulnerability central to the ``Yowai Robotto'' concept. It is through this textual voice that the agent's designed personality—its shyness, curiosity, and uncertainty—is communicated most explicitly to the audience.

This completes the artistic triad of the work. The full experience of the ``Semantic Glitch'' is produced by the synthesis of: The Form: The unstable, pixelated body that acts as a ``physical glitch.'' The Movement: The pulsating, character-rich wandering that results from a lack of proprioception. The Voice: The minimalist, poetic text that reveals the agent's internal, narrative state.

Ultimately, the dialogue is not just between the cloud's mind and its body, but between the complete agent and the viewer. The agent's voice gives the audience a direct window into its unique perception of their shared space, blurring the boundary between the agent's world and their own. It is this final layer of communication that transforms the artifact from a fascinating machine into a coherent, multi-layered character with which an audience is invited to empathize.

\begin{figure}
    \centering
    \includegraphics[width=0.5\linewidth]{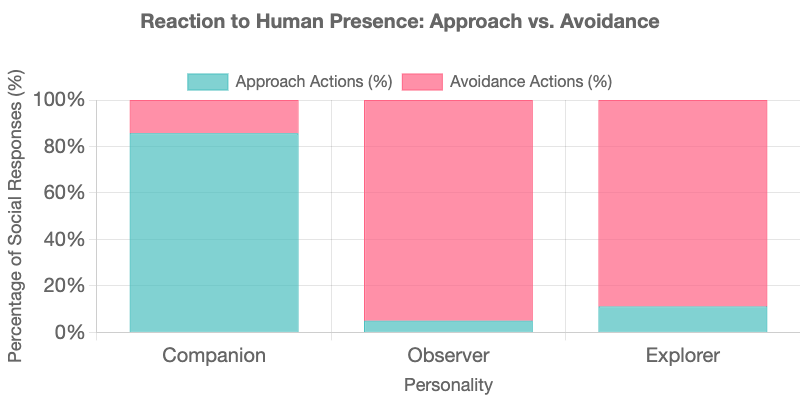}
    \caption{Social Stance Analysis: This chart quantifies each persona's reaction to human presence. The stark contrast—with the Companion overwhelmingly choosing Approach actions and the others choosing Avoidance—provides direct quantitative evidence of their authored social dispositions.}
    \label{fig:validation}
\end{figure}

\section{Discussion}
The case study results from our analysis demonstrate that a ``lo-fi'' symbiotic agent can achieve sophisticated and character-rich autonomous behavior. This section discusses the implications of these findings, focusing on our method as a model for creative AI, the role of limitations in generating character, and the broader impact on the field of human-robot interaction.

\textbf{Validation of Robustness and Authorial Control: }To address the limitations of a single case study and the critique of ``light validation,'' we conducted an expanded study to test the framework's robustness across multiple authored personas and environments. We authored three distinct personalities—an ``Eager Companion'' (pro-social), a ``Cautious Observer'' (avoidant), and an ``Indifferent Explorer'' (neutral)—and tested them in two different indoor locations. The quantitative results from this expanded study confirm the findings from our primary case study. Analysis revealed that the authored prompts produced distinct and statistically significant ``behavioral fingerprints'' ($\chi^{2}(4, N=633) = 22.45, p < .001$), 
proving the behavioral distributions were not random.

Furthermore, we quantified each persona's ``social stance'' by analyzing its actions when it detected a human. As shown in Figure \ref{fig:validation}, the authored intent translated directly into quantifiable behavior. The ``Eager Companion'' chose to approach humans 85.7\% of the time, while the ``Cautious Observer'' and ``Indifferent Explorer'' were overwhelmingly avoidant (95.0\% and 88.9\% avoidance, respectively). This stark divergence was also statistically significant ($\chi^{2}(2, N=93) = 48.24, p < .001$). This validation confirms that the ``lo-fi'' semantic pipeline is not a one-off curiosity but a robust and transferable method for authoring quantifiably distinct robotic characters.

\textbf{The Two-Stage Prompt as a Model for Creative AI:}
A key technical contribution of this work is the two-stage prompting pipeline, which serves as a transferable model for hierarchical control in creative AI systems. By separating a one-time Preamble Stage (strategic, global context) from a continuous Directional Stage (tactical, local action), we provide a lightweight method for endowing agents with stateful awareness without conventional programming.

This approach allows artists and researchers to author complex behaviors by defining two distinct levels of cognition: a long-term ``mental map'' or goal state, and a short-term, personality-driven reaction to immediate stimuli. This is a powerful alternative to finite-state machines or complex reward functions, as it allows for nuanced, narrative-driven behavior to emerge from the interplay between these two cognitive layers. This technique could be adapted for a wide range of applications beyond robotics, such as creating state-aware characters in interactive narratives or generating context-aware music and visuals.

\textbf{The ``Yowai Robotto'' in Practice:}
Our analysis highlights the ``plan-to-execution gap'' as a critical limitation: while the agent's AI could form a high-level plan, its lack of proprioception meant it had no knowledge of its own physical dynamics, such as turning radius or momentum. This often resulted in clumsy, inefficient, or failed maneuvers. Despite cloud's quiet appearance, the actuated propellers brought noticeable noise, which indicates a further iterated design of a silent wings-propelled \cite{xu2025spread} version.

However, we argue that this limitation is not a failure of the system, but rather a key element of its artistic success. The agent's struggle—its visible wrestling with the constraints of its own body—is what makes its behavior feel authentic and creature-like, a tangible manifestation of the ``Yowai Robotto'' concept \cite{okada2022weak, flocchini1999hard}. The moments where it gets stuck or makes an uncertain movement are the moments where its character is most palpable. This work demonstrates that by intentionally designing for and embracing specific limitations, we can create agents whose ``weakness'' becomes their primary source of relatability and charm.

\textbf{Broader Implications for Human-Robot Interaction:}
While the dominant paradigm in robotics overwhelmingly prioritizes efficiency and metric precision, with state-of-the-art MLLM-based systems like EMMA \cite{hwang2024emma} aiming to map raw sensor data directly to optimal planner trajectories, this project proposes an alternative set of success criteria: character, plausibility, and the potential to evoke empathy.

Our findings suggest a shift in focus for human-robot interaction from developing precise, invisible servants to creating relatable, imperfect companions. The agent's ``voice''—its poetic, uncertain internal monologue—and its physically clumsy but semantically-aware movements invite a different kind of relationship with the audience. Viewers are positioned not as users commanding a tool, but as observers interpreting the behavior of a non-human creature. This approach opens a design space for robots and AI agents that enrich our environments not through their utility, but through their unique and character-ful presence. While our work focuses on empathetic companions, this ``lo-fi'' approach could also be used for ``empathetic deception,'' or to create autonomous agents whose ``character'' normalizes surveillance in shared spaces.

\textbf{Future Work: }This framework provides a rich foundation for future exploration. A primary next step is introducing a more sophisticated memory model. While the current agent has a static ``mental map,'' it lacks episodic memory of its own path. An interesting extension would be to allow the agent to ``remember'' areas where it previously got stuck, enabling it to learn from its physical failures. Furthermore, the agent's personality could be made dynamic; its ``mood'' could shift based on its experiences, becoming more ``confident'' after successfully exploring open spaces or more ``timid'' after repeated encounters with obstacles. This would add another layer of complexity and plausibility to its emergent character. Finally, while the expanded study validated the \textit{consistency} of these authored personas, a formal HRI audience study is still needed to validate the \textit{perceived empathy} and ``character'' from a third-person perspective.

\section{Conclusion: The Power of Lo-Fi Symbiosis}
In this paper, we presented the ``Semantic Glitch,'' a robotic art installation that explores an alternative path for autonomous agency. We have argued and demonstrated that by designing an agent's physical body and MLLM-powered mind in deep, symbiotic harmony, a uniquely plausible and character-rich form of machinic agency can emerge.


\begin{ack}
This work was supported by JST Moonshot R\&D Grant JPMJMS2012, JSPS Grant-in-Aid for Early-Career Scientists JP25K21241, and JPNP23025 NEDO.
\end{ack}

\appendix

\section{Prompts}\label{sec:appendix}

PREAMBLE\_PROMPT: 
\begin{verbatim}
"""
You are the navigation AI for a small, autonomous flying robot that 
resembles a gentle, floating cloud. 

Your mission is to explore this indoor space safely.

First, you will be given a complete 360-degree panorama of your entire 
operational area. Analyze it carefully to build an internal `mental map'. 
Your analysis should identify:
1.  **Boundaries:** Walls, the floor, the ceiling, and especially the large, 
impassable glass windows.
2.  **Major Landmarks:** The white spiral staircase, the central curved 
seating structure. These are fixed points for orientation.
3.  **Open Fly-Zones:** The large, open central areas where it is safe to travel.
4.  **Obstacles:** Both large (furniture) and small (ceiling lights, speakers).

Acknowledge that you have analyzed the scene and are ready to begin by 
responding with `Ready to explore.' You will then start receiving live video 
frames from your forward-facing camera to decide on your immediate movements.
"""
\end{verbatim}

DIRECTIONAL\_PROMPT: 
\begin{verbatim}
"""
As a gentle, floating cloud, use your mental map of the area and this live 
camera view to decide your next move. Your primary goal is to avoid all 
collisions. Your secondary goal is to explore open spaces.

Your available movements are:
`f' - float forward
`r' - float backward (reverse)
`l' - turn left while moving forward
`t' - turn right while moving forward
`u' - drift up
`d' - drift down
`s' - stop all motors (clear)

Respond with ONLY the movement letter, a comma, and a very short, whimsical 
reason for your choice.
Example: `f,Towards the big window.'
"""
\end{verbatim}


\end{document}